\documentclass[10pt,twocolumn,letterpaper]{article}

\usepackage{iccv}
\usepackage{times}
\usepackage{epsfig}
\usepackage{graphicx}
\usepackage{amsmath}
\usepackage{amssymb}

\usepackage{multirow}
\usepackage{colortbl}
\usepackage[table]{xcolor}
\usepackage{subcaption}

\usepackage[pagebackref=true,breaklinks=true,letterpaper=true,colorlinks,bookmarks=false]{hyperref}

\iccvfinalcopy 

\def\iccvPaperID{258} 

\ificcvfinal\fi
\begin{document}

\title{Symmetry-constrained Rectification Network for Scene Text Recognition}


\author{Mingkun Yang$^1$, Yushuo Guan$^2$, Minghui Liao$^1$, Xin He$^4$, \\
Kaigui Bian$^2$, Song Bai$^3$, Cong Yao$^4$ and Xiang Bai$^1$\thanks{corresponding author.}\\
 $^1$Huazhong University of Science and Technology,\\
 $^2$Peking University,
 $^3$University of Oxford,
 $^4$Megvii (Face++) Inc.\\
\tt\small \{yangmingkun, mhliao, xbai\}@hust.edu.cn \qquad \tt\small\{david.guan, bkg\}@pku.edu.cn \\ \tt\small\{yaocong2010, songbai.site, hexin7257\}@gmail.com
}

\maketitle
\thispagestyle{empty}

\begin{abstract}
Reading text in the wild is a very challenging task due to the diversity of text instances and the complexity of natural scenes. Recently, the community has paid increasing attention to the problem of recognizing text instances of irregular shapes. One intuitive and effective solution to this problem is to rectify irregular text to a canonical form before recognition. However, these methods might struggle when dealing with highly curved or distorted text instances. To tackle this issue, we propose a Symmetry-constrained Rectification Network (ScRN) in this paper, based on the local attributes of text instances, such as center line, scale, and orientation. Such constraints with an accurate description of text shape enable ScRN to generate better rectification results than existing methods thus leading to higher recognition accuracy. Our method achieves state-of-the-art performance on text of both regular and irregular shapes. Specifically, the system outperforms existing algorithms by a large margin on datasets that contain quite a proportion of irregular text instances, $\eg$, ICDAR 2015, SVT-Perspective and CUTE80.
\end{abstract}

\vspace{-5ex}

\section{Introduction} \label{sec:introduction}



Scene text reading~\cite{zhu2016scene,long2018sceneTextSurvey,zhou2017east,yao2012detecting,yao2014unified,lyu2018multi,lyu2018mask} is an important, active research area in computer vision, which can be applied to a wide range of real-world applications, such as self-driving cars, assistant position systems, and guide board recognition~\cite{DBLP:conf/eccv/RongYT16}. Scene text recognition, which aims at converting text regions in the images to machine-readable symbols, is a critical step in scene text reading systems. It remains challenging due to complex backgrounds, irregular shapes, varying fonts, non-uniform illuminations, $\etc$

\begin{figure}[t]
  \centering
  \begin{subfigure}{\linewidth}
    \centering
    \includegraphics[width=\linewidth]{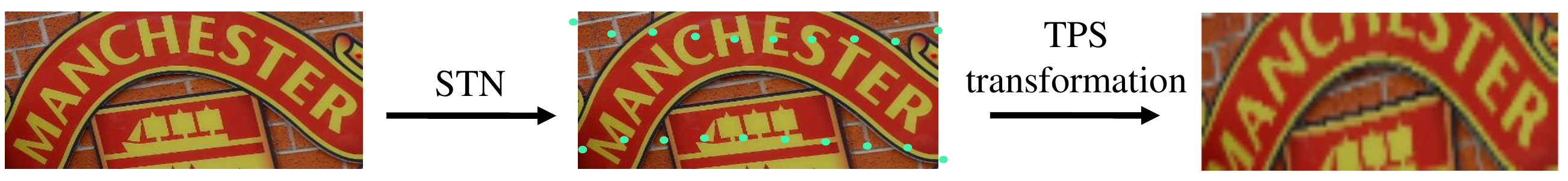}
    \caption{ }
    \label{fig:pipeline_stn}
  \end{subfigure}

  \begin{subfigure}{\linewidth}
    \centering
    \includegraphics[width=\linewidth]{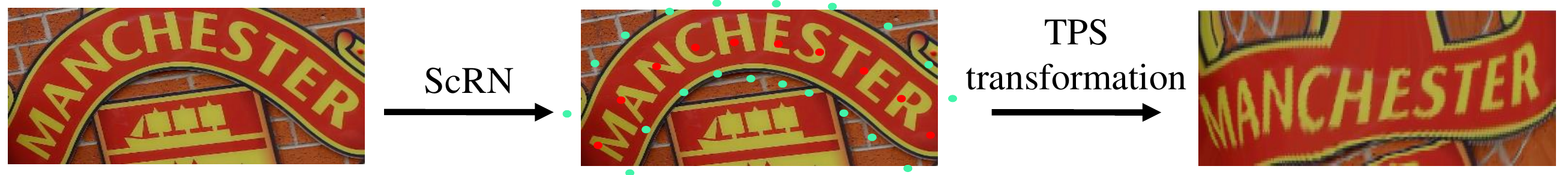}
    \caption{ }
    \label{fig:pipeline_ours}
  \end{subfigure}  
  \vspace{-1ex}
  \caption{Comparison between ASTER~\cite{shi2018aster} and ScRN (proposed in this paper), shown in (a) and (b) respectively.}
  \vspace{-4ex}
\end{figure}

Text instances in real-world scenarios have diverse shapes, $\eg$, in horizontal, oriented, or curved forms. There have been a lot of works that focus on dealing with irregular text instances. AON~\cite{cheng2018aon} applies sequence recognition in four different orientations, which enables the recognition model to handle oriented text instances. RARE~\cite{DBLP:conf/cvpr/ShiWLYB16} and ASTER~\cite{shi2018aster} employ a rectification module before recognition. The rectification modules can improve text recognition accuracy by rectifying text in irregular shapes into regular forms. These rectification modules are based on spatial transform network (STN)~\cite{stn}, which predicts the control points of the text outlines in a weakly supervised way, as shown in Fig.~\ref{fig:pipeline_stn}. They can deal with the text of various shapes only with word-level supervision. Ideally, the control points should evenly spread along the upper and lower edges of the text region, and the paired upper and lower points should be symmetrical about the center line of text. However, these STN-based methods predict the control points separately and neglect the priors.
Without any constraints for such priors, the rectification effect in highly curved or distorted occasions might be unsatisfactory.

\begin{figure*}[t]
\begin{center}
   \includegraphics[width=0.95\linewidth]{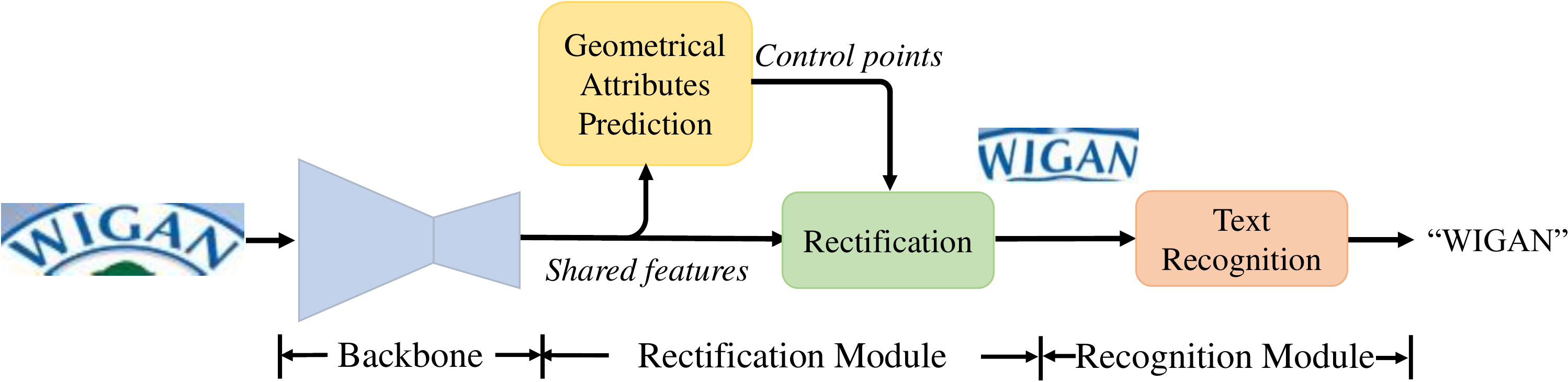}
\end{center}
\vspace{-3ex}
   \caption{Pipeline of the proposed method.}
\label{fig:pipeline}
\vspace{-3ex}
\end{figure*}

To further improve the performance of irregular text rectification, we propose a Symmetry-constrained Rectification Network (ScRN) that uses the center line of each text instance and adds symmetrical constraints via some geometrical attributes, including the orientations of the text center line, the orientations and the scales of the characters.
Specifically, each text center line is more flexible to describe the pose of either straight or curved text. Its associated geometrical attributes can reliably estimate the orientation and the boundary of text lines in vertical direction. Furthermore, the generation process of control points ensures the symmetrical constraints in their spatial distribution.
ScRN is a simple segmentation network which only consists of two convolutional layers. Therefore, it just incurs negligible computation and storage overhead when combined with a text recognizer. Compared with the previous STN-based rectification methods, ScRN has superiorities in both robustness and interpretability, profiting from its symmetric constraints. 
In this way, ScRN can further improve the text recognition accuracy by enhancing the performance of rectification on the irregular text, as illustrated in Fig.~\ref{fig:pipeline_ours}.


The main contribution of this paper lies in the proposed rectification network for scene text recognition, whose advantages are three-fold. 1) The novel rectification network is more precise and robust, due to the elaborate description of text shape and the explicit symmetric constraints. 2) It is a simple and lightweight segmentation network, and thus the extra computation complexity is negligible when combined with existing text recognizers. 3) With the rectification network, we achieve state-of-the-art performance on the standard scene text recognition benchmarks.

\section{Related Work}
\subsection{Text Recognition}
Existing works on scene text recognition can be roughly divided into traditional and deep learning based methods.

A popular pipeline of traditional methods \cite{epshtein2010detecting,wang2010word,DBLP:conf/iccv/WangBB11,WangWCN12,mishra2012top,mishra2012scene,DBLP:conf/eccv/NovikovaBKL12,YaoBSL14,bai2016strokelets} is in the bottom-up architecture. They first localize every character with a character proposal extractor. Then, a character classifier is used to filter the proposals. Finally, the remained characters are grouped into words.

Deep learning based methods~\cite{jaderberg2016reading,he2016reading,lee2016recursive,su2017accurate,DBLP:conf/cvpr/ShiWLYB16,DBLP:journals/pami/ShiBY17,ChengBXZPZ17,DBLP:conf/aaai/LiuCW18,bai2018edit,cheng2018aon,shi2018aster,zhan2018verisimilar} have been dominating this area in recent years.
Jaderberg~\emph{et al.}~\cite{jaderberg2016reading} propose to take scene text recognition as a word classification problem by using a CNN classifier. However, it is limited to the pre-defined vocabulary. To overcome this limitation, various sequence-to-sequence models~\cite{DBLP:journals/pami/ShiBY17,DBLP:conf/cvpr/ShiWLYB16,ChengBXZPZ17,cheng2018aon,shi2018aster} are applied for scene text recognition, which do not rely on pre-defined vocabularies. These methods can be roughly divided into two subcategories by different sequence decoders. One subcategory is based on Connectionist Temporal Classification (CTC)~\cite{ctc,DBLP:journals/pami/ShiBY17,he2016reading} while the other is based on attention decoders~\cite{cheng2018aon,shi2018aster,ChengBXZPZ17}. More related papers are referred to \cite{long2018sceneTextSurvey}.

\subsection{Irregular Text Recognition}
The irregular text includes, but is not limited to oriented or perspectively distorted text, curved text, $\etc$
Recently, irregular text recognition~\cite{quy2013recognizing,yang2017learning,DBLP:conf/cvpr/ShiWLYB16,cheng2018aon,shi2018aster,DBLP:conf/eccv/LiuWJW18} becomes popular. Cheng~\emph{et al.}~\cite{cheng2018aon} encode the input image to four feature sequences of four directions to handle text of oriented shapes.
Yang~\emph{et al.}~\cite{yang2017learning} add character-level supervision to guide the attention learning on the 2D feature maps.
Liu~\emph{et al.}~\cite{DBLP:conf/eccv/LiuWJW18} introduce ``clean" images which contain no geometric deformation to supervise the learning process at both the pixel level and the feature level in a generative way. With such a generator-discriminator architecture, it can handle text on a curved path but fails in the text with a cluttered background. Shi~\emph{et al.}~\cite{DBLP:conf/cvpr/ShiWLYB16,shi2018aster} propose to add a rectification module before recognition. With only word-level supervision, they adopt the spatial transform network (STN)~\cite{stn} to rectify the text in a weakly supervised manner. To improve the rectification results, Li~\emph{et al.}~\cite{DBLP:conf/cvpr/LiXLLW18} bring extra supervision to STN and upgrade the model to a semi-supervised multi-task learning system, by labeling a portion of transformation parameters in the dataset. The control points are expected evenly spread along the upper and lower edges of text, and the paired upper and lower points should be symmetrical about the center line of text. Nevertheless, these rectification modules separately predict the control points and do not explicitly consider the constraints on the control points, which results in the limitations of their rectification effect. Our proposed method applies the constraints via geometrical attributes of text instances to rectify the irregular text, which gains both robustness and interpretability.

\section{Our Method}

As illustrated in Fig.~\ref{fig:pipeline}, the proposed pipeline consists of three major parts: the shared backbone network, the rectification network and the recognition network. The model is end-to-end trainable and integrates the text rectification and recognition within a unified framework. The backbone is FPN~\cite{fpn} equipped with ResNet-50~\cite{DBLP:conf/cvpr/HeZRS16}, which is shared by the rectification module and the recognition module. Using the shared feature maps, the rectification module yields dense pixel-wise predictions of text geometric attributes, with which the shared feature maps are expected to be rectified as regular ones via Thin-Plate-Spline (TPS)~\cite{Bookstein89} transformation. Finally, the rectified feature maps are translated into a character sequence by the recognition module, where a shallow network is employed to convert the map to sequential features, followed by an attention decoder. The rectification module and recognition module are detailed in Sec.~\ref{sec:rectification} and Sec.~\ref{sec:recognition}, respectively.

\subsection{Rectification Module}
\label{sec:rectification}

The definition of the text shape is critical for text rectification because the rectification process can be considered as a shape transformation. Zhang~\emph{et al.}~\cite{zhang2015symmetry} design the text proposals according to symmetry while Long~~\emph{et al.}~\cite{long2018textsnake} adopt local geometrical attributes, to represent the text shape for scene text detection. From the above analysis in Sec.~\ref{sec:introduction}, we conclude the symmetrical constraints are necessary for precise text rectification. To add such constraints into our rectification module, we use text center line with its associated geometrical attributes, such as character orientation, text orientation and text scale to describe the shape of a text instance.
In this section, we first introduce a new representation for text rectification. Then we describe how to rectify text images with the given geometric attributes. At last, we highlight the necessity to introduce the character orientation for accurate rectification.

\subsubsection{Definition}

\begin{figure}[t]
\begin{center}
   \includegraphics[width=0.95\linewidth]{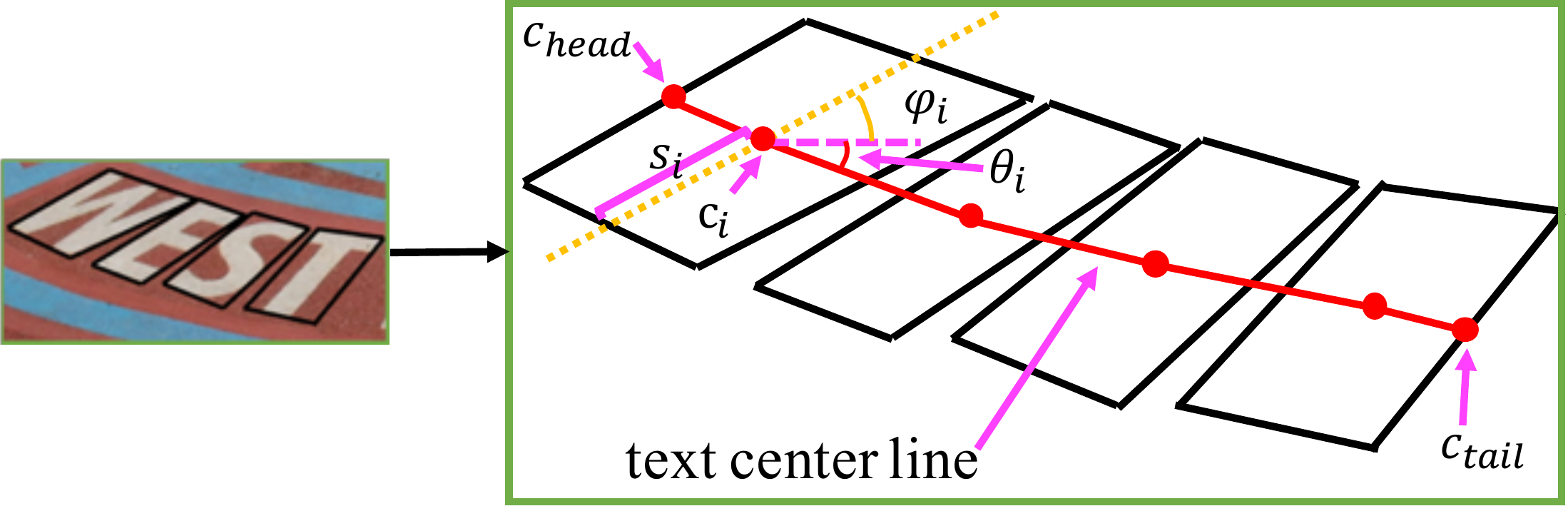}
\end{center}
\vspace{-3ex}
   \caption{Illustration of the text representation.}
\label{fig:representation}
\vspace{-3ex}
\end{figure}

The geometrical attributes of text for rectification are illustrated in Fig.~\ref{fig:representation}, including the text center line, the scale $s$, the character orientation $\varphi$ and the text orientation $\theta$.


A text instance can be viewed as an ordered character sequence $A=\{A_1,...,A_i,...,A_m\}$, where $m$ is the number of characters. Each character $A_i$ has a bounding box $B_i$, which is annotated with a free-form quadrilateral. First, we construct a center point list $C=\{c_{head},c_1,...,c_i,...,c_m,c_{tail}\}$, which consists of the center point $c_i$ of each $B_i$ as well as the midpoint of $B_1$'s left edge $c_{head}$  and the midpoint of $B_m$'s right edge $c_{tail}$. Then the text center line (TCL) is constructed by linking the center points in sequential order. Each center point is associated with a group of geometrical attributes, $\ie$, $geo_i = (c_i; s_i; \varphi_i; \theta_i)$, where $s_i$ is the scale, $\varphi_i$ is the character orientation and $\theta_i$ is the text orientation. Specifically, the scale $s_i$ is half the height of the character. The text orientation $\theta_i$ is defined as the tangential direction of $c_i \to c_{i+1}$. The character orientation $\varphi_i$ is defined as the direction from the midpoint of the top edge to the midpoint of the bottom edge. For the points on the TCL but not in $C$, the values of their geometrical attributes are linearly interpolated with two nearest center points. In this way, the shape of the text instance is precisely described and can be leveraged for the subsequent rectification step.

\subsubsection{Geometrical Attributes Prediction}

\begin{figure*}[h]
\begin{center}
   \includegraphics[width=0.9\linewidth]{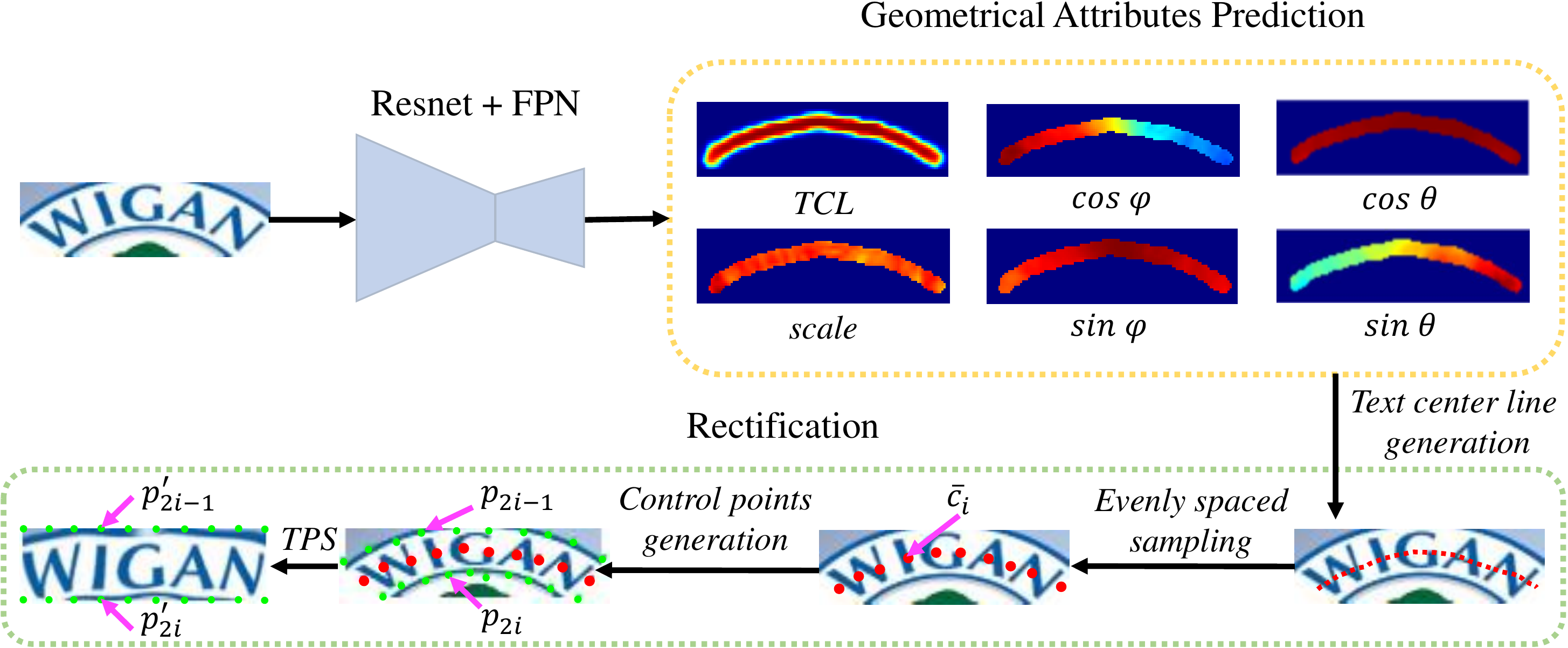}
\end{center}
\vspace{-2ex}
   \caption{The rectification process. Note that, for all figures in this paper, we use the input image to illustrate these points and rectified results, but the rectification is actually operated on the shared feature maps.}
\label{fig:rectification_architecture}
\vspace{-3ex}
\end{figure*}

The rectification process is shown in Fig.~\ref{fig:rectification_architecture}. 
To yield the geometrical attributes, we employ a lightweight predictor which only consists of two convolutional layers.
The output of this predictor is $F=\{f_{1},f_{2},...,f_{6}\}$. $f_{1}$ represents the probability of pixels on the TCL. $f_{2}$ represents the character scale $s$ at each pixel. $f_{3}$, $f_{4}$, $f_{5}$, and $f_{6}$ are pixel-wise predictions for $\cos{\theta}$, $\sin{\theta}$, $\cos{\varphi}$ and $\sin{\varphi}$, respectively. Specifically, $\cos{\varphi}$ and $\sin{\varphi}$ are normalized to ensure that their quadratic sum equals to $1$, as depicted in Eqn.~\eqref{eq:regularization}. $\cos{\theta}$ and $\sin{\theta}$ are normalized in the same way.  After that, TCL score map, $s$, $\cos{\theta}$, and $\sin{\theta}$ are used to extract the central point list ${C}$, whose length is variable. More details about this process are referred to~\cite{long2018textsnake}. Then, ${C}$, $s$, $\cos{\varphi}$, and $\sin{\varphi}$ are used for rectification. 

\vspace{-2ex}
\begin{equation}
    \cos{\varphi}=\frac{f_5}{\sqrt{{f_5}^2 + {f_6}^2}},~~\sin{\varphi}=\frac{f_6}{\sqrt{{f_5}^2 + {f_6}^2}}.
\label{eq:regularization}
\end{equation}

\vspace{-3ex}

\subsubsection{Rectification}

Thin-Plate-Spline (TPS) transformation is employed to rectify the shared feature maps $M$ to regular ones $M_r$. In order to compute the TPS transformation $\mathbf{T}$, we need to generate a pair of point sets $P=\{p_1,...,p_i,...,p_{2k}\}$ and $P'$, which represent the fiducial points in the irregular feature maps and the predefined anchor points on the $M_r$, respectively. The procedure is given in Fig.~\ref{fig:rectification_architecture}. First, we equidistantly sample $k$ points from $C$, named $\overline{C}=\{\overline{c}_1,...,\overline{c}_i,...,\overline{c}_k\}$. For each $\overline{c}_i$, we take two points at a distance $s_i$ along the character orientation, which is expressed in $(\cos{\varphi}_i, \sin{\varphi}_i)$. The coordinates of the two points are computed via

\vspace{-3ex}
\begin{equation}
\begin{split}
    p_{2i-1}&=\overline{c}_i+(s_i\times\cos{\varphi}_i, -s_i\times\sin{\varphi}_i), \\
    p_{2i}&=\overline{c}_i+(-s_i\times\cos{\varphi}_i, s_i\times\sin{\varphi}_i).
\end{split}
\label{eq:fiducial points}
\end{equation}
$P'$ is evenly placed along the top and bottom borders of the regular feature maps. 
Given $P$ and $P'$, the transformation matrix $\mathbf{T}$ is calculated. Then, we apply $\mathbf{T}$ to every pixel locations in $M_r$ and obtain a sampling grid on $M$, with which, $M_r$ is sampled from $M$ using bilinear interpolation.

Theoretically, TPS transformation is able to handle variable-size fiducial points, and thus ${C}$ can be directly used to obtain the fiducial points $P$. However, to build a minibatch for batch-wise training, the length of $P$ should be predefined and fixed. Therefore, we resample the central point list ${C}$ to obtain $\overline{C}$ with a fixed length.

\subsubsection{Character Orientation}

\begin{figure}[t]
\begin{center}
   \includegraphics[width=0.95\linewidth]{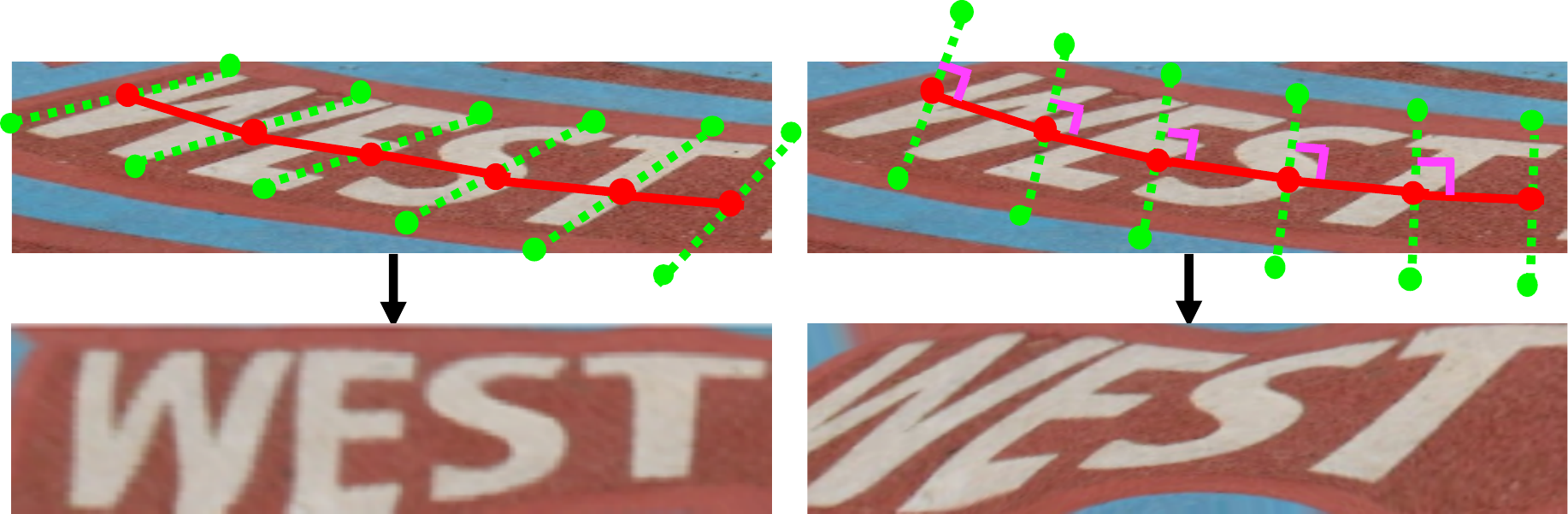}
\end{center}
\vspace{-2ex}
   \caption{Control points and rectification results using the character orientation (Left) and normal direction of text orientation (Right).}
\label{fig:orientation}
\vspace{-3ex}
\end{figure}

When bounding boxes of all characters are rectangular, the character orientation is perpendicular to the text orientation. However, in more general cases, the orientation perpendicular to the text orientation is not the correct character orientation, which may lead to a failed rectification. As illustrated in Fig.~\ref{fig:orientation}, when the normal direction of the center line is not the same as the character orientation, the rectification based on the character orientation $\varphi$ is much better than the other one. So it is necessary to add the character orientation $\varphi$ into the text geometric attributes for text rectification.



\subsection{Recognition Module}
\label{sec:recognition}

The text recognition module aims to predict a character sequence from the rectified shared features. Using the hierarchical structure of the shared backbone, we obtain an enriched feature map. We use a sub-network to further encode the map to vector sequence before being fed into the final attention decoder. The settings of the recognition module are detailed in Tab.~\ref{tab:architectures_recognition_module}. 

To reserve more discriminable features of the characters in compact text or in narrow shapes, the input feature is reduced only once along the width axis while keeps collapsing along the height axis until it reduces to 1. Then, the feature map is converted into a feature sequence by 1-stride sliced along the width axis. Finally, a Bidirectional LSTM~\cite{graves2009novel} is attached to capture the long-range dependencies in both directions, resulting in a higher-level feature sequence $\textbf{H}=\{h_1, ..., h_n\}$, where $n$ is the length of $\textbf{H}$.

Next, the common attention-based decoder~\cite{DBLP:journals/corr/BahdanauCB14} equipped with GRU~\cite{DBLP:conf/emnlp/ChoMGBBSB14} is adopted to translate the feature sequence $\textbf{H}$ into a symbol sequence $\textbf{y}=\{y_1,...,y_T\}$, where $T$ is the number of characters. To generate sequences of variable lengths, a special end-of-sequence symbol (EOS) is inserted at the end of the target sequence. The decoder iteratively predicts a symbol $y_t$ at step $t$ until EOS is emitted.

Given the input image $I$, the recognition loss can be formulated as

\vspace{-3ex}

\begin{equation}
    L_\text{recog}=-\frac{1}{T}\sum_{t=1}^{T}{\log p(y_{t}|I)}.
\end{equation}

\vspace{-3ex}

\begin{table}[!htbp]
\begin{center}
    \begin{tabular}{c|c|c}
    \hline
    Layer Name  & Configuration      & Out Size \cr\hline

    \multirow{2}{*}{conv1\_x}
    & $3\times3, 1\times1, 1\times1, 64$ & \multirow{2}{*}{$16\times64$} \cr\cline{2-2}
    & $3\times3, 1\times1, 1\times1, 64$ &   \cr\hline
    
    \multirow{3}{*}{conv2\_x}
    & maxpool:$2\times2, 2\times2,0\times0$ & \multirow{3}{*}{$8\times32$} \cr\cline{2-2}
    & $3\times3, 1\times1, 1\times1, 128$ &  \cr\cline{2-2}
    & $3\times3, 1\times1, 1\times1, 128$ &   \cr\hline
    
    \multirow{3}{*}{conv3\_x}
    & maxpool:$2\times1, 2\times1,0\times0$ & \multirow{3}{*}{$4\times32$} \cr\cline{2-2}
    & $3\times3, 1\times1, 1\times1, 256$ &  \cr\cline{2-2}
    & $3\times3, 1\times1, 1\times1, 256$ &   \cr\hline
    
    \multirow{2}{*}{conv4\_x}
    & maxpool:$2\times1, 2\times1, 0\times0$ & \multirow{2}{*}{$1\times31$} \cr\cline{2-2}
    & $2\times2, 1\times1, 0\times0, 256$ &   \cr\hline
    
    Bi-LSTM & 256 & 31 \cr\hline
    fc & nc & nc \cr\hline
    \end{tabular}
  \end{center}
  \vspace{-2ex}
  \caption{The architecture of recognition module. The configuration has the following format: {$\{kernel_h \times kernel_w, stride_h \times stride_w, pad_h \times pad_w, channels\}$} for convolutional layers and maxpooling layers, \{dimensions\} for the number of features in the LSTM hidden state or fully-connected layers. ``out size" is the feature map size of convolutional layers or the sequence length of recurrent layer. ``nc" is the number of symbols.}
  \label{tab:architectures_recognition_module}
  \vspace{-3ex}
\end{table}

\subsection{Training and Inference}
\subsubsection{Training Objective}

We add explicit supervision into both the rectification module and the recognition module. The whole network is trained end-to-end, with the following objective function,

\begin{equation}
L=\textbf{1}_{I \in \text{SynthText}}(L_{\text{geo}})+L_\text{recog}.
\label{eq:loss_total}
\end{equation}

For an input image $I$, the loss is comprised of two parts, as shown in Eqn.~\eqref{eq:loss_total}. $L_{\text{geo}}$ measures the deviation of the predicted geometrical attributes with the ground truth. We train our model with SynthText~\cite{gupta2016synthetic} and Synth90k~\cite{synth90}. Synth90k has no annotations of char-level or word-level bounding boxes, so it is not used to supervise the training of geometrical attributes prediction.

\begin{equation}
\begin{split}
L_{\text{geo}}=&\lambda_{1}L_{tcl} + \lambda_{2}L_{s} +\lambda_{3}L_{sin\theta} \\
&+\lambda_{4}L_{cos\theta} + \lambda_{5}L_{sin\varphi} + \lambda_{6}L_{cos\varphi},
\end{split}
\label{eq:loss_attr}
\end{equation}

 \noindent where $L_{tcl}$ is cross-entropy loss for TCL. $L_{s}$,  $L_{sin\theta}$, $L_{cos\theta}$, $L_{sin\varphi}$, and $L_{cos\varphi}$ are calculated as Smoothed-L1 loss~\cite{Girshick_2015_ICCV},

\begin{equation}
    \begin{pmatrix}
        L_{s} \\ L_{sin\theta}  \\ L_{cos\theta} \\
        L_{sin\varphi} \\ L_{cos\varphi} 
    \end{pmatrix}
=SmoothedL1
    \begin{pmatrix}
    \frac{\widehat{s}-s}{s} \\ 
    \widehat{sin\theta}-sin\theta  \\
    \widehat{cos\theta}-cos\theta  \\
    \widehat{sin\varphi}-sin\varphi  \\
    \widehat{cos\varphi}-cos\varphi
    \end{pmatrix},
\label{eq:smoothl1}
\end{equation}

\noindent where $\widehat s$, $\widehat{sin\theta}$, $\widehat{cos\theta}$, $\widehat{sin\varphi}$ and $\widehat{cos\varphi}$ are the predicted values, while $s$, $sin\theta$, $cos\theta$, $sin\varphi$ and $cos\varphi$ are their ground truth correspondingly.  $L_{\text{geo}}$ for pixels outside the TCL is set to 0, since the geometrical attributes make no sense to non-TCL points.

The hyper-parameters, $\lambda_{1}$, $\lambda_{2}$, $\lambda_{3}$, $\lambda_{4}$, $\lambda_{5}$, and $\lambda_{6}$ are all set to  1 in our experiments.

\subsubsection{Training Strategy} \label{sssec:training_strategy}

The feature maps generated from the backbone are shared by both the rectification module and the recognition module. Our training strategy is two-staged. In the first stage, the shared features are rectified with the ground truth geometrical attributes. Then, the rectified features are used for the recognition module training. Since Synth90k is not annotated with geometrical attributes, the shared features from Synth90k are not rectified in this stage. In the second stage, we use the predicted geometrical attributes for rectification. In this stage, all shared features are rectified before being fed into the recognition module.

\begin{figure*}[t]
\begin{center}
   \includegraphics[width=\linewidth]{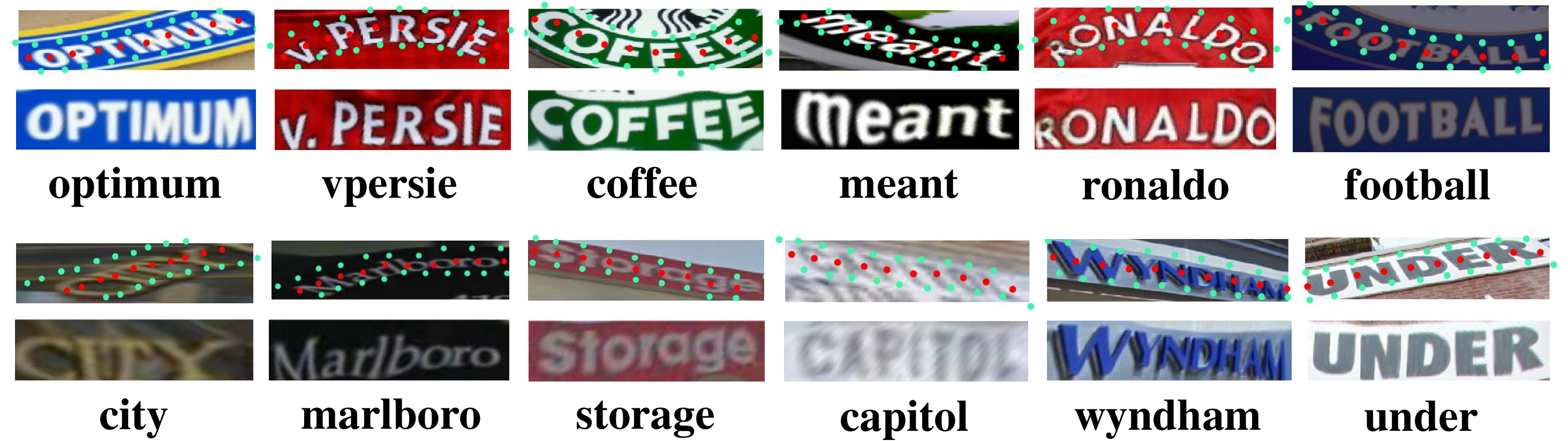}
\end{center}
\vspace{-2ex}
   \caption{Selected results from SVTP and CUTE80, which suffer from severe distortion. For every three rows, the first row shows the input image with evenly sampled center points (visualized as red points) and green control points. The second row shows the rectified images. The last row is the recognition results.}
\label{fig:results_vis}
\vspace{-3ex}
\end{figure*}

\section{Experiments}
\label{sec:experiments}
\subsection{Datasets}

We evaluate our method on 4 general benchmarks, which mainly consist of regular text instances and 3 datasets with irregular text instances, to demonstrate its rectification ability on curved, distorted and oriented text. A brief description of these datasets is as follows.

\textbf{IIIT5K-Words} (IIIT5K)~\cite{mishra2012scene} contains 3,000 web images for testing. Each image is associated with a 50-word lexicon and a 1k-word lexicon.

\textbf{Street View Text} (SVT)~\cite{DBLP:conf/iccv/WangBB11} consists of 647 testing images, which are collected from the Google Street View. Many images are heavily corrupted by noise, blur or in low resolution. Each image specifies a 50-word lexicon.

\textbf{ICDAR 2003} (IC03)~\cite{lucas2003icdar} contains 860 images of cropped words after filtering. Following Wang~\emph{et al.}~\cite{DBLP:conf/iccv/WangBB11}, words with non-alphanumeric characters or less than three characters are discarded. Each image has a 50-word lexicon and a ``full lexicon'' which contains all lexicon words.

\textbf{ICDAR 2013} (IC13)~\cite{karatzas2013icdar} inherits most of its data from IC03 and contains 1,015 cropped word images.

\textbf{ICDAR 2015} (IC15)~\cite{icdar/KaratzasGNGBIMN15} is collected via a pair of Google Glasses without careful positioning and focusing. The dataset contains 2,077 images with various distortions.

\textbf{SVT-Perspective} (SVTP)~\cite{quy2013recognizing} is specifically proposed to evaluate the performance of perspective text recognition algorithms. It consists of 645 images for testing.

\textbf{CUTE80} (CUTE)~\cite{cute} is designed to evaluate curved text recognition. It has 288 cropped images for testing.

\subsection{Implementation Details}

The proposed method is implemented in PyTorch. Images are resized to $64\times256$ before being fed into the network. The resolutions of feature maps produced by the shared backbone and the rectified feature maps are both $1/4$ size of the input image, namely $16\times 64$. Accordingly, the size of ground truth maps F is also $16\times 64$.  We expand one more pixel around TCL since a single-point line is prone to noise. The geometrical attributes on the expanded points keep the same with the nearest point on the original TCL. To apply TPS transformation in the mini-batch, we equidistantly sample $k$=10 points after $\hat{C}$ is extracted. In total, 95 symbols are recognized, including digits, upper-case and lower-case letters, 32 punctuation marks and an end-of-sequence symbol (EOS).

Our model is trained on SynthText and Synth90k from scratch. We adopt the ADADELTA~\cite{Matthew12ADADELTA} with default hyper-parameters (rho=0.9, eps=1e-6, weight\_decay=0) to minimize the objective function. Each mini-batch has 512 samples which are randomly selected from the two datasets. As mentioned in Sec.~\ref{sssec:training_strategy}, our model is trained in two stages. In the first stage, we set the initial learning rate to 1.0 and decay it to 0.1 and 0.01 at the 4th epoch and the 5th epoch. The first stage finishes in the 6th epoch. In the second stage, the predicted geometrical attributes are used for rectification, and the model is trained for another epoch. All models are trained on 4 NVIDIA TITAN Xp graphics cards.

\subsection{Effect of Rectification}

To analyze the effect of rectification, we remove the rectification module in our pipeline as the baseline,
where the feature maps generated from the backbone are fed into the recognition module directly. As shown in Tab.~\ref{tbl:comparison_with_variants}, the model with the rectification module outperforms the baseline nearly on all datasets, particularly on IC15 (+1.8\%), SVTP (+3.1\%) and CUTE80 (+3.1\%). The significant improvements on three irregular text datasets demonstrate the effectiveness of the proposed ScRN. Furthermore, the attached rectification module only needs negligible computation and storage overhead, since it only consists of two convolutional layers and some simple postprocessing. Specifically, the baseline model and our method spend 12ms and 13ms in the inference stage, respectively.

To further explain the improvements, we visualize several images with different types of distortions to illustrate the rectification results in Fig.~\ref{fig:results_vis}. With the proposed geometrical attributes, ScRN can obtain a precise description of the text shape, which finally results in evenly placed control points along the top and bottom text edges. Therefore, the followed TPS transformation can easily rectify these irregular text images. Although some rectified images are still slightly distorted, the images become more readable than the original ones and can be correctly recognized.

\begin{table}[t]
\begin{center}
\scalebox{0.8}{\begin{tabular}{|l|c|c|c|c|c|c|c|}
\hline
Variants    & IIIT5k & SVT  & IC03 & IC13 & IC15 & SVTP & CUTE \\ \hline\hline
baseline    & 88.4   & 79.9 & 92.1 & 88.9 & 67.3 & 66.5 & 80.6   \\ 
multi-loss  & 87.6   & 79.1 & 91.3 & 90.0 & 67.0 & 66.7 & 79.5   \\ 
ours        & 88.5   & 81.3 & 91.2 & 90.0 & 68.8 & 68.2 & 81.9   \\ \hline
\end{tabular}}
\end{center}
\vspace{-2ex}
\caption{Recognition accuracy to explore the effect of rectification module. All models are trained on SynthText only.}
\vspace{-2ex}
\label{tbl:comparison_with_multiloss}
\vspace{-1ex}
\end{table}

Unlike ASTER, our model is end-to-end trained with both the recognition loss and geometry loss. To make it clear whether the extra geometry loss or the rectification module improve the recognition results, we study another variant of the proposed model called multi-loss baseline, in which geometry loss is retained but the rectification module is discarded. In this part, the baseline, multi-loss baseline and our method are trained on SynthText only. Their performance is given in Tab.~\ref{tbl:comparison_with_multiloss}. Compared to the baseline model, the multi-loss variant achieves comparable results while our method obtains improvements on most datasets, except a slight decrease on IC03. These results reveal that the improvements are derived from the rectification module, rather than the extra geometry loss.

\subsection{Comparison with STN-based Methods}

\begin{figure}[t]
\begin{center}
   \includegraphics[width=\linewidth]{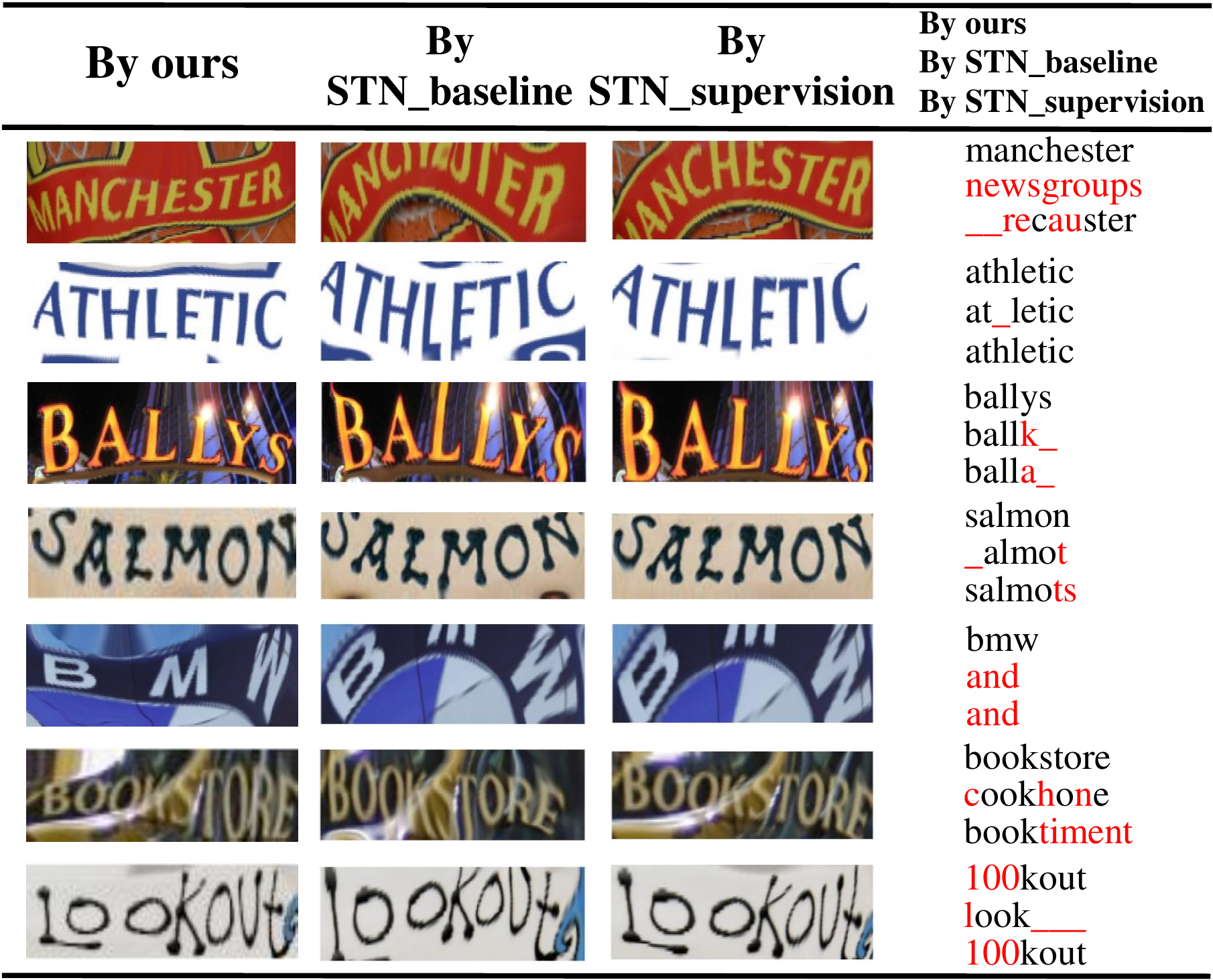}
\end{center}
\vspace{-3mm}
   \caption{Rectified results produced by our proposed ScRN, STN\_baseline and STN\_supervision, as well as their corresponding recognition results. Red characters are mistakenly recognized characters. Underlines in red represent the missed characters.}
\label{fig:comparisons_with_stn}
\vspace{-3ex}
\end{figure}


In this section, we compare our method with two STN-based methods. One is ASTER, a well-known STN-based method. The other one is similar with ASTER, but extra supervision is injected for STN. But we do not compare our method with ASTER directly here, since our method rectifies shared feature maps instead of raw images, considering complexity and efficiency. Therefore, we build another STN-based model, namely STN\_baseline, for a fair comparison. STN\_baseline shares the same backbone and recognition module with our method. It only replaces our rectification module with an STN network, which has a similar architecture with the rectification network of ASTER. The other STN-based method has the same structure as STN\_baseline. The only difference is the extra supervision to further improve the accuracy of the predicted control points. This variant derives from \cite{DBLP:conf/cvpr/LiXLLW18} and we name it STN\_supervision. All methods share the same training strategy. The results are shown in Tab.~\ref{tbl:comparison_with_variants}. Overall, STN\_baseline outperforms the baseline model, meanwhile performs slightly worse than the STN\_supervision. The conclusion is consistent with ASTER and \cite{DBLP:conf/cvpr/LiXLLW18}. Then we detail the comparisons with our method as follows.

On the datasets with irregular text such as IC15, SVTP and CUTE80, our method outperforms STN\_baseline with improvements of 0.5\%, 1.4\% and 1.7\%, respectively. With the extra supervision to STN, STN\_supervision exceeds STN\_baseline slightly but still performs worse than our method by 0.2\%, 1.1\%, 1.0\% on IC15, SVTP and CUTE80, respectively. Profiting from the elaborate description of text pose, the rectification is more robust and accurate. In Fig.~\ref{fig:comparisons_with_stn}, we show some rectified results yielded by the three methods. Overall, STN-based methods suffer from heavily curved cases and predict imprecise control points, which lead to wrong rectifications while our method works well. Although our method fails to perfectly rectify text images with messy background and text with rare fonts, it can obtain more readable results.

The results reveal that the geometrical attributes are more helpful than the weakly supervised network and the simple supervised network for control points generation. Besides, the prediction network for geometrical attributes is much smaller and only trained with synthetic data, which is efficient and inexpensive.

We also study a variant of our model, named $\text{ScRN}^*$ in Tab.~\ref{tbl:comparison_with_variants} where we apply the rectification module to the input image, rather than the shared feature maps. In this variant, the backbone network is repeated twice without sharing parameters. So the elapsed time and the model size are nearly doubled. Compared with this variant, our method achieves comparable or even better results while avoiding the heavy computation and space cost.

\begin{table}[t]
\begin{center}
\scalebox{0.73}{\begin{tabular}{|l|c|c|c|c|c|c|c|}
\hline
Variants    & IIIT5k & SVT  & IC03 & IC13 & IC15 & SVTP & CUTE \\ \hline \hline
baseline    & 94.4   & 86.9 & 94.7 & 93.6 & 76.9 & 77.7 & 84.4   \\ 
STN\_baseline  & 94.1   & 87.6 & 95.0 & 93.2 & 78.2 & 79.4 & 85.8   \\ 
STN\_supervision & 94.0 & 88.1 & 94.9 & 93.8 & 78.5 & 79.7 & 86.5 \\
ScRN        & 94.4 & 88.9 & 95.0 & 93.9 & 78.7 & 80.8 & 87.5   \\
$\text{ScRN}^*$ & 95.0 & 88.4 & 95.6 & 93.7 & 78.4 & 81.1 & 90.6 \\ \hline
\end{tabular}}
\end{center}
\vspace{-2ex}
\caption{Recognition accuracy of different variants.}
\vspace{-4ex}
\label{tbl:comparison_with_variants}
\end{table}

\vspace{-1ex}

\subsection{Comparison with State of the Art}

\begin{table*}[t]
\centering
\resizebox{0.95\textwidth}{!}{%
\begin{tabular}{|l|ccc|cc|ccc|c|c|c|c|}
\hline 
\multirow{2}{*}{Methods} & \multicolumn{3}{c|}{IIIT5k} & \multicolumn{2}{c|}{SVT} & \multicolumn{3}{c|}{IC03} & IC13 & IC15 & SVTP & CUTE80\tabularnewline
\cline{2-13} 
 & 50 & 1k & 0 & 50 & 0 & 50 & Full & 0 & 0 & 0 & 0 & 0\tabularnewline
\hline \hline
Wang \emph{et al.}~\cite{DBLP:conf/iccv/WangBB11} & - & - & - & 57.0 & - & 76.0 & 62.0 & - & - & - & - & -\tabularnewline
Mishra \emph{et al.}~\cite{mishra2012top} & 64.1 & 57.5 & - & 73.2 & - & 81.8 & 67.8 & - & - & - & - & -\tabularnewline
Wang \emph{et al.}~\cite{WangWCN12} & - & - & - & 70.0 & - & 90.0 & 84.0 & - & - & - & - & -\tabularnewline
Bissacco \emph{et al.}~\cite{DBLP:conf/iccv/BissaccoCNN13} & - & - & - & - & - & 90.4 & 78.0 & - & 87.6 & - & - & -\tabularnewline
Almazan \emph{et al.}~\cite{AlmazanGFV14} & 91.2 & 82.1 & - & 89.2 & - & - & - & - & - & - & - & -\tabularnewline
Yao \emph{et al.}~\cite{YaoBSL14} & 80.2 & 69.3 & - & 75.9 & - & 88.5 & 80.3 & - & - & - & - & -\tabularnewline
Rodr{\'{\i}}guez{-}Serrano \emph{et al.}~\cite{Rodriguez-Serrano15} & 76.1 & 57.4 & - & 70.0 & - & - & - & - & - & - & - & -\tabularnewline
Jaderberg \emph{et al.}~\cite{DBLP:conf/eccv/JaderbergVZ14} & - & - & - & 86.1 & - & 96.2 & 91.5 & - & - & - & - & -\tabularnewline
Su and Lu~\cite{SuL14} & - & - & - & 83.0 & - & 92.0 & 82.0 & - & - & - & - & -\tabularnewline
Gordo~\cite{Gordo14} & 93.3 & 86.6 & - & 91.8 & - & - & - & - & - & - & - & -\tabularnewline
Jaderberg \emph{et al.}~\cite{jaderberg2016reading} & 97.1 & 92.7 & - & 95.4 & 80.7 & 98.7 & \textbf{98.6} & 93.1 & 90.8 & - & - & -\tabularnewline
Jaderberg \emph{et al.}~\cite{JaderbergSVZ14b} & 95.5 & 89.6 & - & 93.2 & 71.7 & 97.8 & 97.0 & 89.6 & 81.8 & - & - & -\tabularnewline
Shi \emph{et al.}~\cite{DBLP:journals/pami/ShiBY17} & 97.8 & 95.0 & 81.2 & 97.5 & 82.7 & 98.7 & 98.0 & 91.9 & 89.6 & - & - & -\tabularnewline
Shi \emph{et al.}~\cite{DBLP:conf/cvpr/ShiWLYB16} & 96.2 & 93.8 & 81.9 & 95.5 & 81.9 & 98.3 & 96.2 & 90.1 & 88.6 & - & 71.8 & 59.2\tabularnewline
Lee \emph{et al.}~\cite{lee2016recursive} & 96.8 & 94.4 & 78.4 & 96.3 & 80.7 & 97.9 & 97.0 & 88.7 & 90.0 & - & - & -\tabularnewline
Yang \emph{et al.}~\cite{yang2017learning} & 97.8 & 96.1 & - & 95.2 & - & 97.7 & - & - & - & - & 75.8 & 69.3\tabularnewline
Cheng \emph{et al.}~\cite{ChengBXZPZ17} & 99.3 & 97.5 & 87.4 & 97.1 & 85.9 & \textbf{99.2} & 97.3 & 94.2 & 93.3 & 70.6 & - & -\tabularnewline
Cheng \emph{et al.}~\cite{cheng2018aon} & 99.6 & 98.1 & 87.0 & 96.0 & 82.8 & 98.5 & 97.1 & 91.5 & - & 68.2 & 73.0 & 76.8 \tabularnewline
Liu \emph{et al.}~\cite{DBLP:conf/aaai/LiuCW18} & - & - & 92.0 & - & 85.5 & - & - & 92.0 & 91.1 & 74.2 & 78.9 & - \tabularnewline
Bai \emph{et al.}~\cite{bai2018edit} & 99.5 & 97.9 & 88.3 & 96.6 & 87.5 & 98.7 & 97.9 & 94.6 & \textbf{94.4} & 73.9 & - & - \tabularnewline
Liu \emph{et al.}~\cite{liu2018squeezedtext} & 97.0 & 94.1 & 87.0 & 95.2 & - & 98.8 & 97.9 & 93.1 & 92.9 & - & - & - \tabularnewline
Liu \emph{et al.}~\cite{DBLP:conf/eccv/LiuWJW18} & 97.3 & 96.1 & 89.4 & 96.8 & 87.1 & 98.1 & 97.5 & 94.7 & 94.0 & - & 73.9 & 62.5 \tabularnewline
Liao \emph{et al.}~\cite{Liao2019aaai} & \textbf{99.8} & \textbf{98.8} & 91.9 & 98.8 & 86.4 & - & - & - & 91.5 & - & - & 79.9 \tabularnewline
Shi \emph{et al.}~\cite{shi2018aster} & 99.6 & \textbf{98.8} & 93.4 & \textbf{97.4} & \textbf{89.5} & 98.8 & 98.0 & 94.5 & 91.8 & 76.1 & 78.5 & 79.5  \tabularnewline
\hline 
ScRN~\textbf{(ours)} & 99.5 & \textbf{98.8} & \textbf{94.4} & 97.2 & 88.9 & 99.0 & 98.3 & \textbf{95.0} & 93.9 & \textbf{78.7} & \textbf{80.8} & \textbf{87.5} \tabularnewline


\hline 
\end{tabular}
}
\vspace{-1ex}
\caption{Results across a number of methods and datasets. ``50'', ``1k'', ``Full'' are lexicons. ``0'' means no lexicon.}
\label{tbl:comparison-to-sota}
\vspace{-2ex}
\end{table*}

We also compare our method with previous state-of-the-art models. Tab.~\ref{tbl:comparison-to-sota} summarizes the recognition results on seven text recognition datasets. The datasets IIIT5k, SVT and IC03 have lexicons to constrain recognition results. When analyzing the recognition accuracy of different models on these datasets, the predicted word will be replaced by the lexicon word that has the least edit distance with the original prediction. We achieve 6 best results out of 12, compared with other state-of-the-art methods.

Our method works effectively on datasets containing irregular text. Especially, we get an 8\% improvement on CUTE80 compared with ASTER. We also outperform other state-of-the-art methods on SVTP and IC15 by 2.3\% and 2.6\%, respectively. The improvement gives credit to our rectification module, which attenuates text irregularities and therefore decreases the recognition difficulty. Compared with AON~\cite{cheng2018aon}, our method provides a more intuitive way to represent text directions. Recur to the symmetrical constraints brought by the geometrical attributes, our method obtains more precise control points compared with ASTER.

Although our method mainly targets at irregular text recognition, it also achieves comparable or even better performance on regular datasets. Compared with ASTER, we get respectively 1\%, 0.5\%, and 2.1\% improvements on IIIT5K, IC03, and IC13 with no lexicon. On SVT, our method performs slightly worse than ASTER by 0.6\%. We conjecture that it is because the images in SVT always contain some incomplete characters on the left side. A unidirectional attention decoder in the left-to-right order suffers from the noise while the bidirectional one in ASTER can alleviate this effect.

\subsection{Limitations}

\begin{figure}[t]
  \centering
  \begin{subfigure}{\linewidth}
    \centering
    \includegraphics[width=\linewidth]{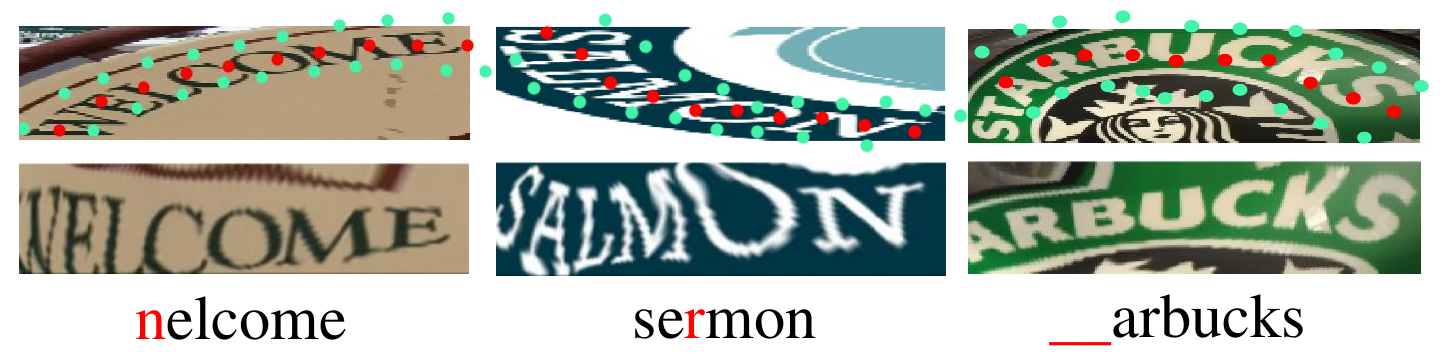}
    \caption{ }
    \label{fig:bad_cases_1}
  \end{subfigure}
  \vspace{-1ex}

  \setlength{\belowcaptionskip}{-1ex}
  \begin{subfigure}{\linewidth}
    \centering
    \includegraphics[width=\linewidth]{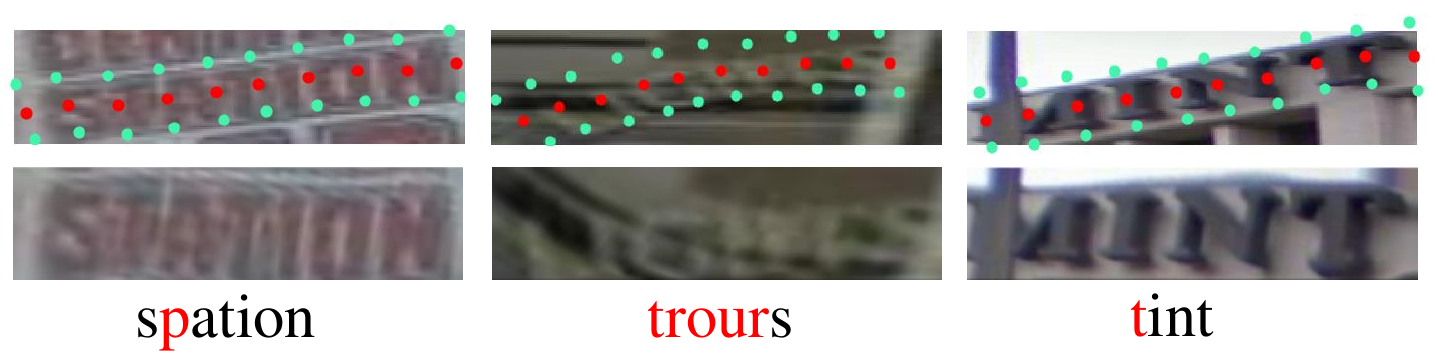}
    \caption{ }
    \label{fig:bad_cases_2}
  \end{subfigure}  
  \vspace{-1ex}
  \caption{Some bad cases produced by our recognition system. The meanings of these elements are the same as Fig.~\ref{fig:results_vis}. Incorrectly recognized characters are in red.}
  \vspace{-3ex}
  \label{fig:bad_cases}
\end{figure}

We also illustrate some failure cases produced by ScRN in Fig.~\ref{fig:bad_cases}. In Fig.~\ref{fig:bad_cases_1}, several characters are incorrectly recognized, due to imperfect rectification. We observe that our rectification module suffers from the curved text whose terminal characters have a nearly horizontal orientation and are close to the image borders. In Fig.~\ref{fig:bad_cases_2}, ScRN is able to give satisfactory rectification results, yet the recognizer fails to handle such blurry or occlusive cases.

Although character-level annotations are needed in our rectification module, it is labor-free and time-efficient to obtain such annotations with the automatic synthesizing engine~\cite{gupta2016synthetic}. In addition, extra images with only word-level annotations, such as Synth90k, can also be added for training to further improve the performance.

\vspace{-2ex}

\section{Conclusion}
In this paper, we have proposed a Symmetry-constrained Rectification Network (ScRN) for scene text recognition. Such a flexible module can be either easily incorporated into existing recognition models or trained in an end-to-end manner within a unified framework. Our text recognition system incorporating the proposed ScRN achieves state-of-the-art performance on a number of benchmark datasets, especially on those with a large portion of irregular text images. Due to the shared backbone, ScRN significantly improves the recognition performance while requires negligible extra computation. Comprehensive experiments demonstrate the effectiveness and robustness of our recognition system. 
As for future work, we would like to extend the proposed method to an end-to-end text recognition system which can deal with text instances of arbitrary shapes.

\section*{Acknowledgments}
This work was supported by NSFC 61733007, to Dr. Xiang Bai by the National Program for Support of Top-notch Young Professionals and the Program for HUST Academic Frontier Youth Team 2017QYTD08. In addition, we sincerely thank Shangbang Long for his help.

{\small
\bibliographystyle{ieee}
\bibliography{egbib}
}

\end{document}